\documentclass[journal]{IEEEtran}
\usepackage{amsmath,amsfonts,bm}
\usepackage{amssymb}
\usepackage{algpseudocode}
\usepackage[ruled,linesnumbered]{algorithm2e}
\usepackage{array}
\usepackage{textcomp}
\usepackage{stfloats}
\usepackage{url}
\usepackage{verbatim}
\usepackage{graphicx}
\usepackage{balance}
\usepackage{color,soul}
\usepackage[ruled]{algorithm2e}
\usepackage{caption}
\usepackage{subcaption}
\usepackage[numbers,sort&compress]{natbib}

\ifCLASSINFOpdf
\else
\fi

\makeatletter
\newcommand{\removelatexerror}{\let\@latex@error\@gobble}
\makeatother

\begin{document}
\title{GraphGANFed: A Federated Generative Framework for Graph-Structured Molecules Towards Efficient Drug Discovery}

\author{
Daniel~Manu, 
Jingjing~Yao,~\IEEEmembership{Member, IEEE}, 
Wuji~Liu,
and~Xiang Sun,~\IEEEmembership{Member, IEEE}, 
\thanks{D. Manu and X. Sun are with the SECNet Lab., Department of Electrical and Computer Engineering, University of
New Mexico, Albuquerque, NM 87131, USA. E-mail: $\{$dmanu,sunxiang$\}$@unm.edu.\par
J. Yao is with the Department of Computer Science, Texas Tech University, Lubbock, TX 79409, USA. E-mail: jingjing.yao@ttu.edu.\par
W. Liu is with Amazon, San Diego, CA, 92121, USA. E-mail: liuwujicoding@gmail.com.\par
This work was supported by the National Science Foundation under Award CNS-2148178.
}
}

\maketitle

\begin{abstract}
Recent advances in deep learning have accelerated its use in various applications, such as cellular image analysis and molecular discovery. In molecular discovery, a generative adversarial network (GAN), which comprises a discriminator to distinguish generated molecules from existing molecules and a generator to generate new molecules, is one of the premier technologies due to its ability to learn from a large molecular data set efficiently and generate novel molecules that preserve similar properties. However, different pharmaceutical companies may be unwilling or unable to share their local data sets due to the geo-distributed and sensitive nature of molecular data sets, making it impossible to train GANs in a centralized manner. In this paper, we propose a \ul{Graph} convolutional network in \ul{G}enerative \ul{A}dversarial \ul{N}etworks via \ul{Fed}erated learning (GraphGANFed) framework, which integrates graph convolutional neural Network (GCN), GAN, and federated learning (FL) as a whole system to generate novel molecules without sharing local data sets. In GraphGANFed, the discriminator is implemented as a GCN to better capture features from molecules represented as molecular graphs, and FL is used to train both the discriminator and generator in a distributive manner to preserve data privacy. Extensive simulations are conducted based on the three benchmark data sets to demonstrate the feasibility and effectiveness of GraphGANFed. The molecules generated by GraphGANFed can achieve high novelty $\left(\approx 100 \right)$ and diversity $\left(> 0.9\right)$. The simulation results also indicate that 1) a lower complexity discriminator model can better avoid mode collapse for a smaller data set, 2) there is a tradeoff among different evaluation metrics, and 3) having the right dropout ratio of the generator and discriminator can avoid mode collapse. 

\end{abstract}


\section{Introduction}
The discovery of new organic and inorganic molecules remains a challenge in medicine, chemistry, and materials sciences. Traditional approaches to molecular discovery involve mathematical frameworks derived from related properties calculated from chemical structures with different physical or biological reactions \cite{nouira,walters2020}. However, these mathematical frameworks may not fully capture the properties of the chemical structures, limiting the ability to fully explore novel molecules. To address this challenge, computer-aided solutions have been developed that utilize deep learning methods to learn highly complex representations of chemical structures, extract features from existing molecules, and generate new molecules based on the learned features. These approaches have shown promise in facilitating molecular generation \cite{manu2021a,guimaraes2017,bagal2021,cae2018} and \emph{de novo} drug design \cite{denovo2021}. 

\begin{figure}[t]
    \centering
    \captionsetup{justification=centering}
    \includegraphics[width=7cm, height=6.5cm]{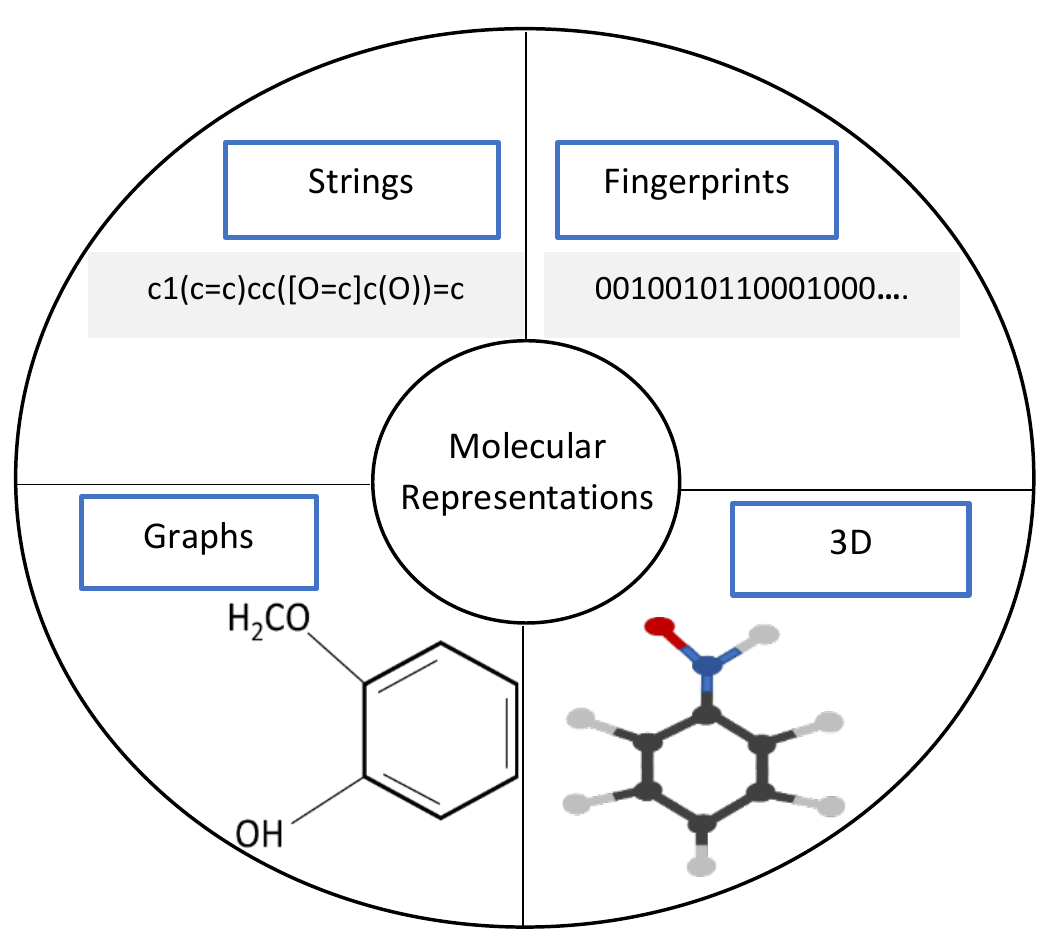}
    \caption{Different molecular representations.}
    \label{fig:mol_rep}
\end{figure}

Deep generative models, one of the deep learning models widely applied in molecular discovery, generate molecules based on features/patterns learned from the existing molecule sets with desired chemical properties. There are two common types of input data for generative models: Simplified Molecular Input Line Entry System (SMILES) representations \cite{smiles1988} and molecular graphs \cite{gaudelet2021}. SMILES is a string-based representation, has been applied as the input data type for Natural Language Processing (NLP) deep learning models such as Recurrent Neural Networks (RNNs) \cite{rnn2017} and Auto-Encoders \cite{simonvsky2018,jin2018,lim2018,putin2018,poly2018}. Recent works have explored the use of RNN-based generative models to learn the properties of the existing molecules represented as SMILES, and then generate novel molecules with the same properties. Auto-Encoder based models consist of an encoder and a decoder, where the encoder encodes the molecules into a latent vector representation and the decoder transforms the latent vectors back to molecules. In contrast to SMILES, molecular graphs use graphs in terms of edges and nodes to represent the structures of molecules. Molecular graphs can represent more detailed structures of molecules than SMILES, thus attracting much attention \cite{simonvsky2018}. Fig. \ref{fig:mol_rep} shows different molecular representations for generative modeling. 

Graph Convolutional Neural Network (GCN) is a typical deep learning model that can learn molecular graph representations and analyze the related properties ~\cite{graph2019,graph2018,manu2021b}. For example, \cite{twostep} applies GCN to estimate a specific metric (e.g., novelty, validity, or uniqueness) of a molecule by analyzing the structure of the molecule represented as a molecular graph. 
Generative Adversarial Networks (GANs) have emerged as another popular approach for molecular discovery \cite{nouira,manu2021a,guimaraes2017}. A GAN model consists of a generator and a discriminator that compete with each other. The generator generates realistic new samples that preserve the features of the existing samples to deceive the discriminator, while the discriminator tries to differentiate between the generated and existing samples. The generator and discriminator are trained iteratively, and the generator becomes better at generating realistic samples that deceive the discriminator over time. GANs have been widely used in various image generation tasks and have demonstrated outstanding performance \cite{bao2017,han2018,andreini2020}. 
The performance of GANs for molecular generation can be further improved by integrating GCNs since GCNs can learn more detailed properties of molecules represented as molecular graphs. Hence, the integration of GCN and GAN in drug discovery is worth investigation. 


To derive an accurate deep learning model for generating new molecules, it is essential to train the model on a large and diverse data set. However, it is not realistic to obtain this data set because the sharing of different pharmaceutical companies' molecular databases is not always feasible due to the compound intellectual property and strict data regulation \cite{chen2021}. One of the existing solutions is that pharmaceutical companies encrypt their data samples and share the encrypted data samples with a third party, which trains the model over the encrypted data samples by using Secure Multiparty Computation (MPC) \cite{mpc2018}. The encryption method, however, cannot fully ensure data privacy, thus hindering collaborative drug discovery. 
Recently, Federated Learning (FL) has emerged as a promising solution to train deep learning models in a distributed and privacy-preserving manner \cite{9417084}. In FL, a centralized FL server broadcasts a global model to all the clients in terms of pharmaceutical companies. Each client trains its received global model over its local data set for several epochs and then uploads its trained local model to the FL server. After that, the FL server aggregates the received local models from the clients to generate a new global model, and then broadcasts the new global model to the clients to start the next global round. This process continues until the global model is converged. The clients only share their local models and do not share their local data sets, which ensures data privacy. Hence, FL can enable pharmaceutical companies to collaborate and train accurate deep learning models on their collective data while preserving data privacy, which is important to accelerate the drug discovery process. 

In this work, we propose the \ul{Graph} convolutional network in \ul{G}enerative \ul{A}dversarial \ul{N}etworks via \ul{Fed}erated learning (GraphGANFed) framework for molecular discovery, which leverages FL to train a GAN model that uses GCNs to represent molecules as graphs. In this framework, the discriminator is implemented by a GCN that learns the features/properties of the existing molecules, while the generator is implemented by a multilayer perceptron (MLP) or a multilayer neural network that generates new molecules while preserving the properties of the existing molecules to deceive the discriminator. The discriminator and generator are trained based on FL, which ensures data privacy as each client keeps its data locally. In GraphGANFed, the discriminator evaluates the molecules generated by the generator. This evaluation guides the generator to generate new molecules that preserve features of existing molecules but have different structures, so that the generator can deceive the discriminator later on. Hence, it is critical to design a good discriminator model that can perform an accurate evaluation to guide the generator in generating more realistic new molecules. 
Therefore, we investigate how different discriminator model structures affect the performance metrics (e.g., Validity, Novelty, Diversity, etc.) in our framework GraphGANFed. We also analyze how different dropout ratios affect the evaluation metrics. The main contributions of our work are summarized as follows.

\begin{itemize}
    \item We propose a GraphGANFed framework that combines GCN, GAN, and FL to generate novel and effective molecules while preserving data privacy. 
    \item We implement our GraphGANFed framework and demonstrate its performance based on three benchmark datasets. The results indicate that GraphGANFed generates molecules with high novelty $\left(\approx 100 \right)$ and diversity $\left(> 0.9\right)$.
    \item We investigate the impact of various factors on the performance of GraphGANFed, including the complexity of the discriminator, number of clients, and dropout ratios via extensive simulations.
\end{itemize}

The rest of the paper is organized as follows. Section II presents a brief background and related works regarding molecular discovery, GAN, and GCN. Section III introduces the design of the GraphGANFed architecture, and the metrics to evaluate generated molecules. Section IV discusses the simulation setups and results. Finally, Section V concludes the paper. 

\section{Background and related works}
In this section, we provide a brief overview of GANs, state-of-the-art generative models for molecular discovery, challenges and proposed solutions in GANs, and GCN frameworks for molecular discovery.

\subsection{Basis of GANs}
GANs are generative models that aim to minimize the divergence between the real data distribution $P_d$ and the generative model distribution $P_z$. Given a training data distribution $P_d$, the generator generates samples $x$ from the distribution $P_z$ with the random noise $z$. The discriminator discriminates between real samples from $P_d$ and generated samples from $P_z$ \cite{goodfellow2020}. The generator aims to minimize the loss $\log (1-D(\Tilde{x}))$, where $\Tilde{x}$  is the generator's output when given noise $z$ and $D(\Tilde{x})$ is is the discriminator's estimate of the probability that a generated sample is an existing sample. On the other hand, the discriminator tries to maximize the probability of correctly labeling both real and generated samples. Overall, GANs can be viewed as a two-player min-max game: 
\begin{equation}
    \min_{\theta}\max_{\phi}{\mathbb{E}}_{x\sim P_d}[log D_{\phi}(x)]+{\mathbb{E}}_{\Tilde{x}\sim P_{z}}[log(1-D_{\phi}(G_\theta (z))],
    \label{e:gan}
\end{equation}
where $D_{\phi}(\cdot)$ and $G_\theta (\cdot)$ are the discriminator and generator models parameterized by $\phi$ and $\theta$, respectively. ${\mathbb{E}}_{x\sim P_d}[log D_{\phi}(x)]$ is the expected value over all the existing samples for all the probabilities that the discriminator correctly classifies the existing samples and ${\mathbb{E}}_{\Tilde{x}\sim P_{z}}[log(1-D_{\phi}(G_\theta (z))]$ is the expected value over all the generated samples for all the probabilities that the discriminator correctly classifies the generated samples from the generator.  

\subsection{Generative models for molecular discovery}
GANs have emerged as a highly promising approach for generating novel and diverse molecules, and have become the most commonly used generative models for state-of-the-art molecular generations. Several recent studies have demonstrated the power of GANs in the field. Guimaraes \textit{et al.} \cite{guimaraes2017} proposed an objective-reinforced generative adversarial network (ORGAN) by integrating reinforcement learning (RL) into the GAN. They used the SeqGAN (sequence based Generative Adversarial Network) architecture based on RNNs and GAN, and applied a policy gradient method to learn the generative network parameters. 
Sanchez-Lengeling \textit{et al.} \cite{organic2017} proposed the objective-Reinforced GAN for Inverse-Design Chemistry (ORGANIC), a modified ORGAN framework, to design novel chemical compounds with desired properties. ORGANIC learns a biased generative distribution that shifts the output toward regions in the molecular space that optimizes desired chemical properties. ORGANIC was demonstrated to generate novel molecules with high scores of desired chemical properties. Putin \textit{et al.} \cite{putin2018} proposed a deep neural network Adversarial Threshold Neural Computer (ATNC) architecture. This architecture is based on GAN and RL, where the generator is implemented as a Differential Neural Computer (DNC) with new specific blocks known as adversarial threshold (AT). Their framework successfully generated good molecules whilst overcoming the negative reward problem in ORGANIC.

\subsection{Challenges and solutions in GANs}
GANs are widely known for generating expected samples but they suffer from the mode collapse issue (i.e., generate limited variety of samples) and may be difficult to train. Hence, several studies have explored methods to stabilize GAN training. Arjovsky \emph{et al.} \cite{wass2017} proposed a Wasserstein distance $W(P_{z}, P_{d})$ to measure the distance between the existing data distribution $P_{d}$ and generated data distribution $P_z$. Under soft conditions (i.e., generator is locally Lipschitz and continuous in $\theta$), $W(P_{z}, P_{d})$ is continuous and differentiable almost everywhere. 
By minimizing the Wasserstein distance, a GAN can generate more realistic samples and better approximate the existing data distribution. 

The min-max game for Wasserstein GAN (WGAN) can be expressed by the Kantorovich-Rubinstein duality \cite{optim2009}, i.e.,
\begin{equation}\small
    \min_{\theta}\max_{\phi\in\mathcal{D}}{\mathbb{E}}_{x\sim P_d}[D_{\phi}(x)]-{\mathbb{E}}_{\Tilde{x}\sim P_{z}}[D_{\phi}(\Tilde{x})],
    \label{e:wgan}
\end{equation}
where $\mathcal{D}$ is the set of 1-Lipschitz functions, $P_{z}$ is the model distribution implicitly defined by $\Tilde{x}=G(z)$, and $z\sim p_{z}$ is a random noise vector generated from a prior distribution $p(z)$. Hence, for an optimal discriminator (which is also called a critic since it is not to classify samples, but to evaluate the Wassertein distance between generated samples and existing samples), reducing the value function with respect to the generator parameters, i.e., from Eq. \eqref{e:wgan}, reduces $W(P_d, P_z)$. 

To ensure that the discriminator satisfies the Lipschitz constraint, they proposed weight clipping of the critic within a dense space $[-c, c]$. However, they demonstrated that the weight clipping approach leads to optimization problems in \cite{improvedwass2017}. 
To address the stability, vanishing, and exploding gradient issues, Gulrajani \textit{et al.} \cite{improvedwass2017} proposed an alternative approach to enforce the Lipschitz constraint. They introduced a soft constraint with a penalty on the gradient norm for random samples $\hat{x}\sim {P}_{\hat{x}}$, where $P_{\hat{x}}$ is the distribution of random samples from the generated and existing data distributions. Thus, the new loss function of the discriminator becomes
\begin{equation}\footnotesize
    L = \underset{\text{original critic loss}}{\underbrace{-{\mathbb{E}}_{x\sim P_d}[D_{\phi}(x)]+{\mathbb{E}}_{\widetilde{x}\sim P_{z}}[D_{\phi}(\widetilde{x})]}}+\underset{\text{gradient penalty}}{\underbrace{{\gamma{\mathbb{E}}_{\hat{x}\sim P_{\hat{x}}}[(\lVert{\nabla_{\hat{x}}D(\hat{x})\rVert_{2}-1)^2]}}}},
    \label{e:wgan1}
\end{equation}
where $\gamma$ is the gradient penalty coefficient. In our work, we adopt the loss function in Eq. \eqref{e:wgan1} to stabilize the training process and mitigate model collapse.

\subsection{Graph Convolutional Networks} 

GCNs have been applied in various drug discovery applications such as biological property \cite{mol2016}, quantum mechanical property \cite{quan2017}, interaction prediction \cite{padme2018,drug2018}, synthesis prediction \cite{pred2017}, and \emph{de novo} molecular design \cite{simonvsky2018}. In \cite{mol2016}, a graph convolutional framework was proposed to learn the molecular representations using both node and edge features. Each layer is comprised of atoms (nodes) in a molecule and pair-wise representations. The framework uses a Weave unit to establish relations between different layers through propagation between atoms to atoms, atoms to pairs, pairs to atoms, and pairs to pairs. Every layer adopts the weave unit architecture, except for the last convolution layer where only the atom representations are utilized for down-streaming applications, such as estimating solubility or drug efficacy. The framework uses neural networks to model the transitions over the same representations (i.e., atom-atom and pair-pair). However, for the transitions over different representations (i.e., atom-pair and pair-atom), an additional order-invariant aggregation operation is used for the feature transition. The inputs of the graph convolutional framework are molecular graphs containing atoms and bond features. The authors evaluated their proposed framework for biological activities on 259 data sets consisting of PubChem BioAssay (PCBA) \cite{pubchem2012}, the training set for the Tox21 challenge \cite{deeptox2016}, the enhanced directory of useful decoys \cite{decoys2012}, and the ``maximum unbiased validation" data sets built by Rohrer and Baumann \cite{muv2009}, in a multi-task setting where biological activities are predicted. However, the results show that the proposed framework did not always outperform the baseline methods that used Morgan fingerprints.

In \cite{simonvsky2018}, the authors proposed a framework named GraphVAE for molecular generations, which directly generates a probabilistic fully-connected graph with a predefined maximum size. GraphVAE is based on the variational autoencoder method, where the stochastic graph encoder is used to process discrete attributed graphs $G=(A,E,F)$, representing the adjacency matrix $A$, edge attribute tensor $E$, and node attribute tensor $F$ with a predefined number of nodes. The encoder is implemented by a feed-forward network with edge-conditioned graph convolutions. The stochastic graph encoder embeds the graphs into a continuous representation $z$. The graph decoder outputs a probabilistic fully-connected graph on the predefined nodes, where the discrete samples can be drawn. To ensure that the output graph matches the ground truth, they used a standard matching algorithm to align the output graph to the ground truth. GraphVAE was evaluated based on two molecular data sets, QM9 and ZINC, and the results demonstrated that it can learn appropriate embedding properties for small molecules, but it is difficult to capture complex chemical interactions for larger molecules.  

All the above existing works about molecular generation and drug discovery are performed in a centralized setting which results in mode collapse and hinders the ability of the models to generalize on diverse datasets. Our proposed framework aims to solve this model collapse and diversity issues in the centralized setting by ensuring that models are trained in a federated setting without sharing the client dataset thus preserving data privacy.   

\section{Graph Convolutional Network in Generative Adversarial Networks via Federated learning (GraphGANFed)}
In this section, we explain the key concepts in molecular graphs, provide the detailed design of our proposed GraphGANFed architecture, and analyze the metrics to evaluate the performance of our proposed GraphGANFed architecture.

\subsection{Molecular graphs} \label{sec:Molecular_Graphs}
In GraphGANFed, every molecule is represented as a non-directional graph $\mathcal{G}$ with a set of nodes $\mathcal{V}$ and a set of edges $\mathcal{E}$. Each atom in the molecule corresponds to a node in the graph, and the type of each atom is specified using a node label matrix $\bm{V}$. The matrix $\bm{V}=\left\{ v_{ij}\left| 1 \le i\le N, \right. 1 \le j \le B \right\}$, where $N$ is the total number of atoms in a molecule, $B$ is the number of possible atom types, and $v_{ij}$ implies if atom $i$ is the type $j$ (i.e., $v_{ij}$=1) or not (i.e., $v_{ij}$=0). Here, we consider generating molecular graphs with the fixed maximum number of atoms $N=10$. Each atom belongs to one of the 10 possible types (i.e., $B=10$): Carbon (C), Nitrogen (N), Oxygen (O), Fluorine (F), Bromine (Br), Phosphorus (P), Sulfur (S), Chlorine (Cl), Iodine (I), and one padding symbol (*). The padding symbol is used to represent an atom other than the types mentioned above, and it allows for flexibility in the molecular structure to accommodate the individual design preferences of different clients. 

In GraphGANFed, the bonds between atoms in a molecule are represented as non-directional edges $(i, i')\in \mathcal{E}$, where $i, i' \in \mathcal{V}$. An adjacency matrix $\bm{A}$ is used to depict all the bonds between the atoms in a molecule. The matrix $\bm{A}=\left\{ a_{ii'k}\left| 1\le i,i'\le N,1\le k\le T\right. \right\}$, where $T$ is the number of bond types, and $a_{ii'k}$ implies if the bond between atoms $i$ and $i'$ is bond type $k$ (i.e., $a_{ii'k}=1$) or not (i.e., $a_{ii'k}=0$). Here, we consider 5 types of bonds (i.e., $T=5$): zero, single, double, triple, and aromatic bonds. 

\begin{figure}[t]
    \centering
    \captionsetup{justification=centering}
    \includegraphics[width=9cm, height=5.5cm]{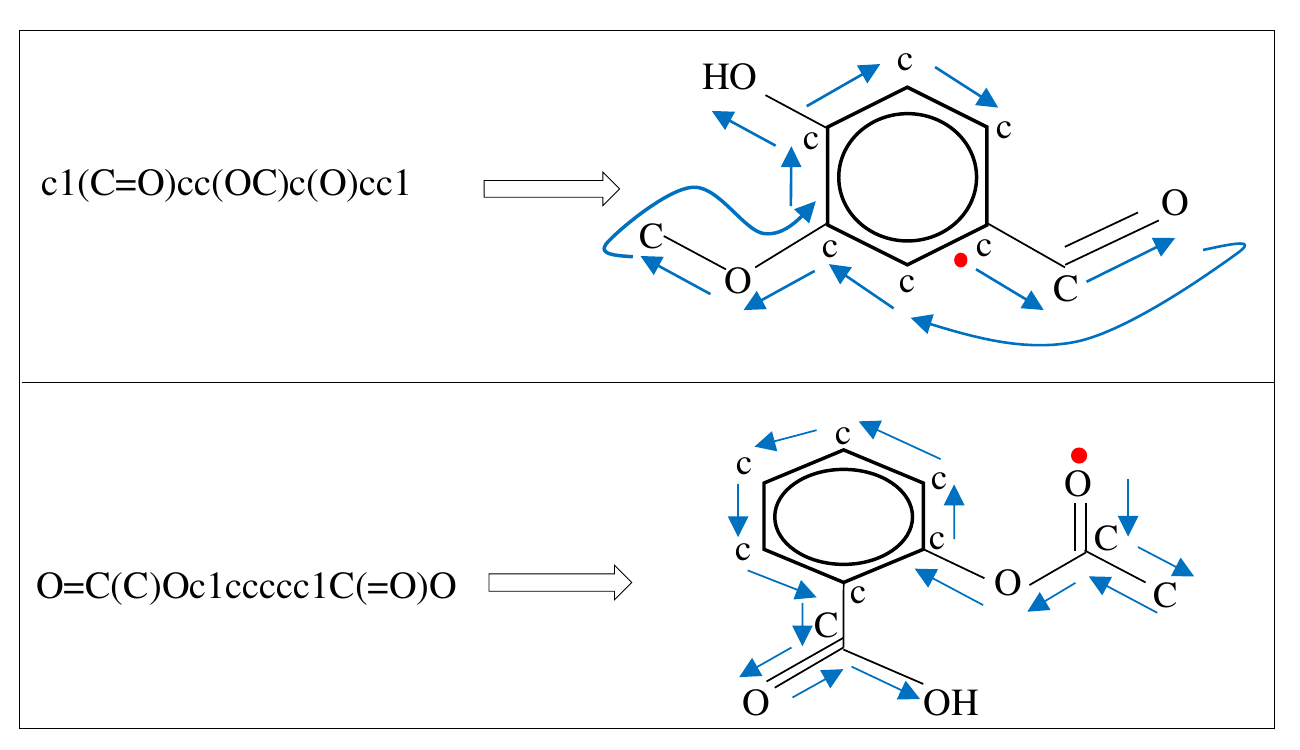}
    \caption{Examples of converting SMILES to molecular graphs.}
    \label{fig:smiles2graph}
\end{figure}

\subsection{The GraphGANFed framework} 
\begin{figure*}[!htb]
	\centering	
    \includegraphics[width=1.0\textwidth]{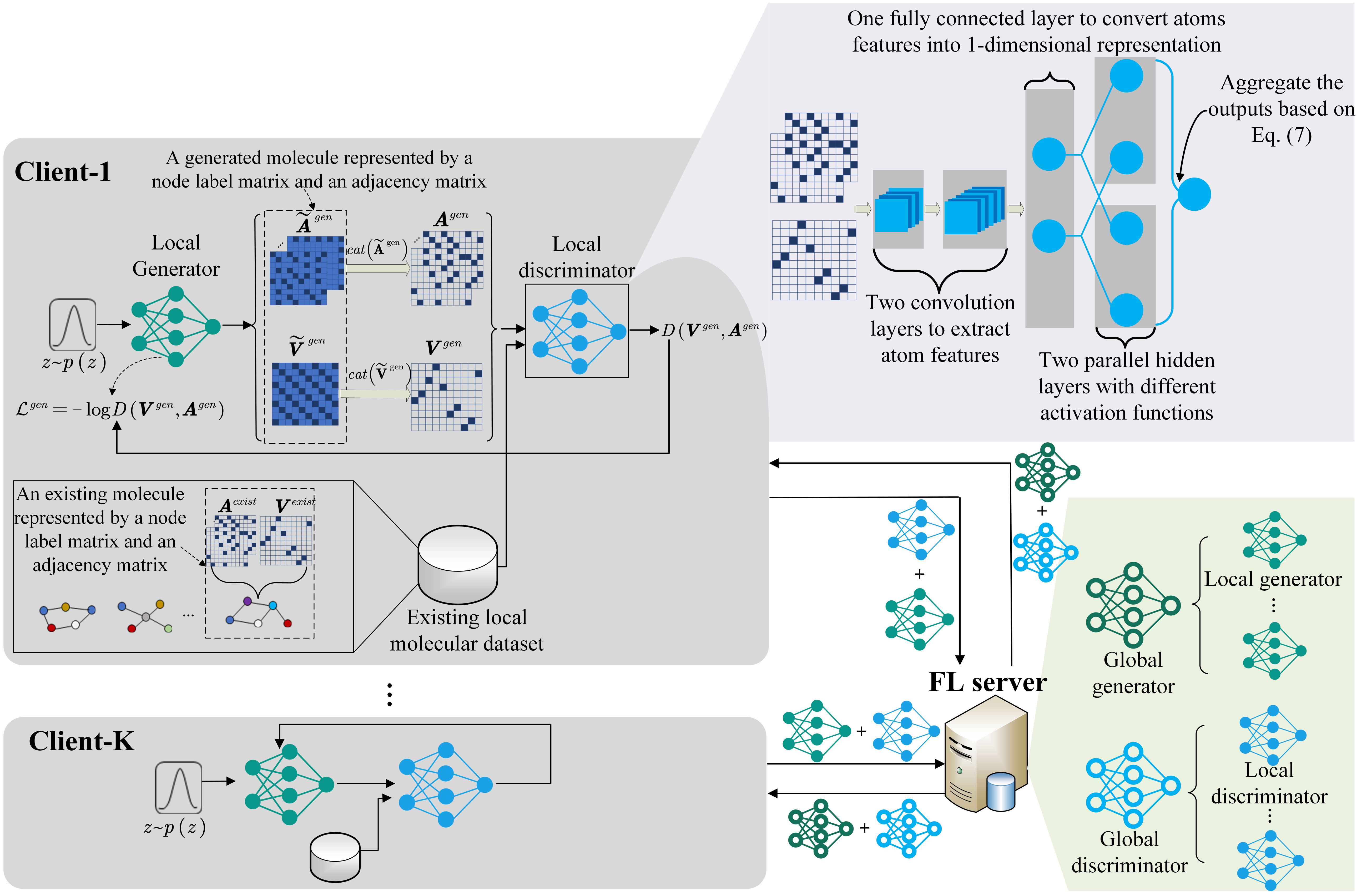} 
	\caption{Overview of the GraphGANFed framework.}
	\label{fig:over} 
\end{figure*}

In general, GraphGANFed consists of three main parts, i.e., the generator, discriminator, and FL system. Fig. \ref{fig:over} shows the overview of GraphGANFed. 

\begin{figure}
    \centering
    \vspace{-50pt}
    \includegraphics[width=\columnwidth]{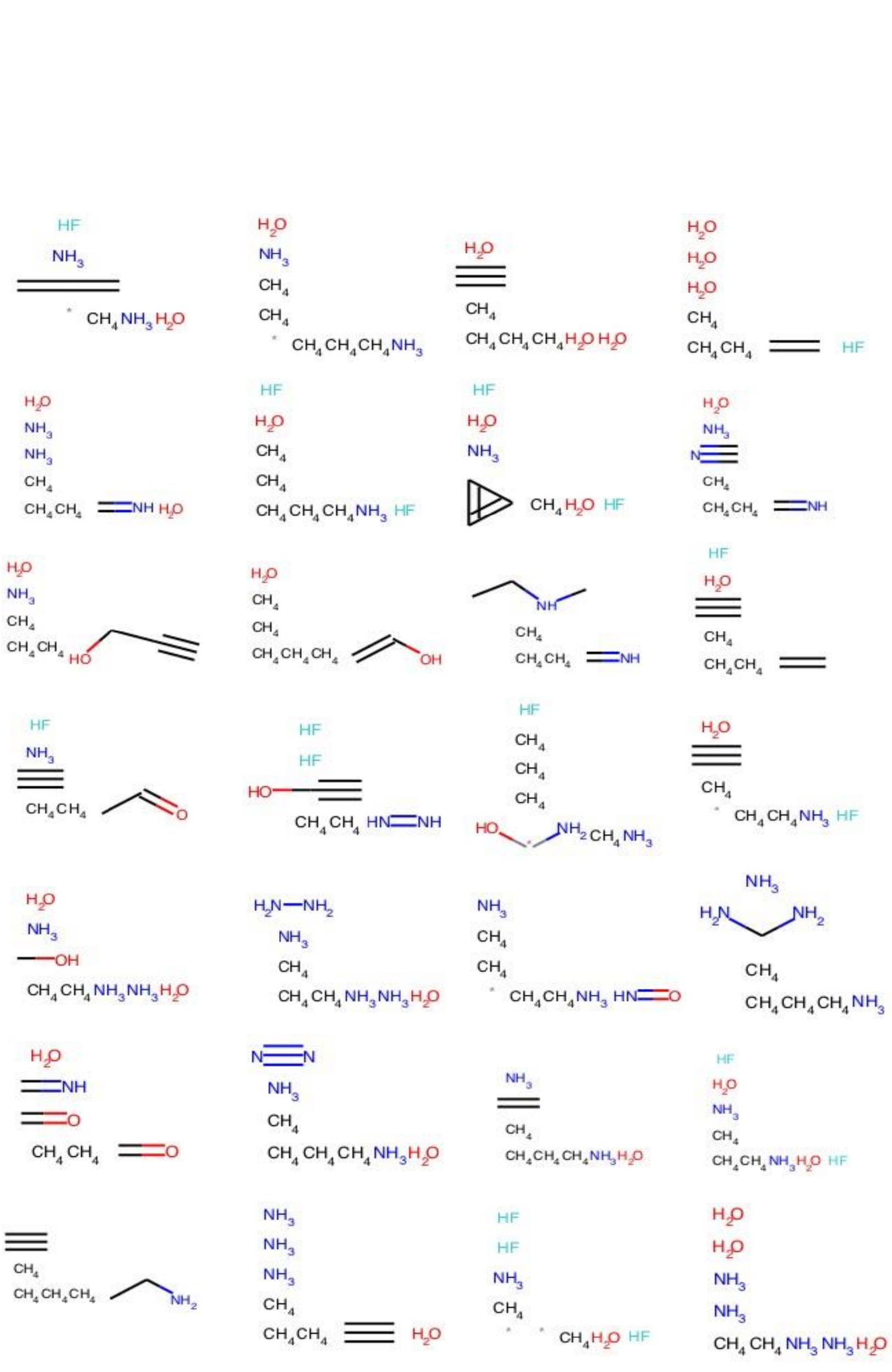}
    \caption{Examples of molecules generated by GraphGANFed.}
    \label{fig:mols}
\end{figure}

\textbf{The Generator} is implemented using a multi-layer perceptron (MLP) neural network. It takes a random noise vector $z$, which is sampled from a prior distribution $p(z)$, as an input and outputs the molecular graph represented by two continuous matrices, i.e., node label matrix $\bm{V}$ and adjacency matrix $\bm{A}$. Each hidden layer in the MLP neural network applies the tanh activation function. The output of the final hidden layer of the MLP is linearly projected to match the dimensions of $\bm{V}$ and $\bm{A}$, and then normalized using the softmax function. Fig. \ref{fig:mols} shows some molecules generated by the described MLP neural network. The goal of the generator is to minimize the following loss function \cite{molgan2018}
\begin{equation}
    \mathcal{L}^{gen}=-\log D(\bm{\Tilde{V}}^{gen}, \bm{\Tilde{A}}^{gen}),
\end{equation}
where $\bm{\Tilde{V}}^{gen}$ and $\bm{\Tilde{A}}^{gen}$ are the node label and adjacency matrices of a generated molecule, respectively, and $D(\bm{\Tilde{V}}^{gen}, \bm{\Tilde{A}}^{gen})$ is the output of the discriminator for the generated molecule. 

It is worth noting that both the generated and existing molecules will be used to train the discriminator, which will be explained later on. However, the node label and adjacency matrices of a generated molecule (i.e., $\bm{\Tilde{V}}^{gen}$ and $\bm{\Tilde{A}}^{gen}$) are continuous, while the node label and adjacency matrices of an existing molecule (denoted as $\bm{V}^{exit}$ and $\bm{A}^{exit}$) are discrete, as they are derived based on the definition of molecular graphs described in Section \ref{sec:Molecular_Graphs}. To make the types of matrices consistent (so that they can be fed into the same discriminator model for training), we discretize $\bm{\Tilde{V}}^{gen}$ and $\bm{\Tilde{A}}^{gen}$ by applying categorical sampling \cite{gumbel2016}. Specifically, we obtain the discrete node label and adjacency matrices of a generated molecule by applying categorical sampling as follows: $\bm{V}^{gen}=\mathrm{cat}(\bm{\Tilde{V}}^{gen})$ and $\bm{A}^{gen}=\mathrm{cat}(\bm{\Tilde{A}}^{gen})$, where $\bm{V}^{gen}$ and $\bm{A}^{gen}$. 

\textbf{The Discriminator} aims to distinguish the molecules generated by the generator from the those in the existing data set, and so the loss function of the discriminator is \cite{improvedwass2017}
\begin{align}
     \mathcal{L}^{dis}=&-\log D(\bm{V}^{gen}, \bm{A}^{gen})+\log D(\bm{V}^{exist}, \bm{A}^{exist}) \nonumber\\
     &+\gamma(||\nabla_{\bm{\Omega_{A}},\bm{\Omega_{V}}}D(\bm{\Omega_{A}},\bm{\Omega_{V}})||-1)^2,
\label{eq:loss_disc}
\end{align}
where $\bm{V}^{exist}$ and $\bm{A}^{exist}$ are the node label and adjacency matrices of a molecule in the existing data set, respectively, $D(\bm{V}^{exist}, \bm{A}^{exist})$ is the output of the discriminator for the existing molecule, and $\gamma(||\nabla_{\bm{\Omega_{A}},\bm{\Omega_{V}}}D(\bm{\Omega_{A}},\bm{\Omega_{V}})||-1)^2$ is the gradient penalty term to stabilize the gradients of the discriminator. Here, $\gamma$ is the gradient penalty coefficient, $\bm{\Omega_{A}}=\varepsilon \bm{A}^{exist}+(1-\varepsilon)\bm{A}^{gen}$, $\bm{\Omega_{V}}=\varepsilon \bm{V}^{exist}+(1-\varepsilon)\bm{V}^{gen}$, and $\varepsilon \in [0,1]$ is a predefined hyperparameter. The intuition of the loss function $\mathcal{L}^{dis}$ is to maximize the difference between the output of an existing molecule and the output of a generated molecule i.e., $D(\bm{V}^{exist}, \bm{A}^{exist})-D(\bm{V}^{gen}, \bm{A}^{gen})$. A larger difference implies the two molecules can be easily discriminated. Note that the two outputs $D(\bm{V}^{exist}, \bm{A}^{exist})$ and $D(\bm{V}^{gen}, \bm{A}^{gen})$ are positive scalar values capturing the features of the molecules. 

The discriminator is implemented using a GCN, which convolves the node label matrix $\bm{V}$ using the adjacency matrix $\bm{A}$. Here, we use the relational GCN (R-GCN) encoder, which is a convolutional network that supports non-directional and multiple-edge graphs as inputs. The discriminator consists of two convolution layers to extract the atom features, followed by one layer MLP to compute a one-dimensional feature representation, which will be parallelly fed into two pieces of one hidden layer, as illustrated in Fig. \ref{fig:over}. The final output layer is used to aggregate the features from the two hidden layers followed by the tanh activation function to generate a scalar value representing the features of the input molecule. Specifically, the atom feature representations are propagated through each of the first two layers based on \cite{manu2021a,molgan2018} 
\begin{equation}
\begin{aligned}
    h_{i}^{'(l+1)} & = f_{s}^{(l)}(h_{i}^{(l)}, \bm{v}_i)+\sum_{i'=1}^{N}\sum_{k=1}^{T}\frac{a_{ii'k}}{\lvert \mathcal{N}_i \rvert}f_{t}^{(l)}(h_{i'}^{(l)}, \bm{v}_i), \\
    h_{i}^{(l+1)} & = \mathrm{tanh}(h_{i}^{'(l+1)}),
    \label{e:prop}
\end{aligned}
\end{equation}
where $h_{i}^{(l)}$ is the feature of atom $i$ at layer $l$, $\bm{v_i}$ is the type of atom $i$ (i.e., the $i^\text{th}$ row in the node label matrix $\bm{V}$), $f_{s}^{(l)}$ is the residual or skip connection between layer $l$ and the next layer $l+1$, $f_{t}^{(l)}$ is an edge-type affine function of layer $l$, ${\mathcal{N}}_{i}$ is the set of atom $i$'s neighboring atoms in the molecule\footnote{Here, we define two atoms in a molecule are the neighbors if there is an edge (i.e., single, double, triple, or aromatic bond) between the two atoms.}, $\left| {\mathcal{N}}_{i} \right|$ implies the number of neighbors for atom $i$. After propagating the node feature representations through 2 convolution layers, the output is converted into a 1-dimensional graph representation, denoted as $h_i^{(L)}$, which is simultaneously fed into two parallel hidden layers. Denote $\sigma(g_1(h_{i}^{(L)}, \bm{v}_{i}))$ and $tanh(g_2(h_{i}^{(L)}, \bm{v}_{i}))$ as the outputs of the two parallel hidden layers after applying the sigmoid and tanh activation functions, respectively, where $g_1(h_{i}^{(L)}, \bm{v}_{i})$ and $g_2(h_{i}^{(L)}, \bm{v}_{i})$ are the outputs of the two parallel hidden layers before their activation functions. The final output layer will aggregate the results from the two parallel hidden layers by conducting element-wise multiplication followed by the tanh activation function, i.e.,
\begin{equation}
\begin{aligned}
    h_{\mathcal{G}}^{'} & =\sum_{i=1}^{N}\sigma(g_1(h_{i}^{(L)}, \bm{v}_{i}))\odot \mathrm{tanh}(g_2(h_{i}^{(L)}, \bm{v}_{i})), \\
    h_{\mathcal{G}} & = \mathrm{tanh}(h_{\mathcal{G}}^{'}),
    \label{e:agg}
\end{aligned}
\end{equation}
where $h_{\mathcal{G}}$ and $h'_{\mathcal{G}}$ are the results before and after the tanh activation function, and $\odot$ is element-wise multiplication. Here, $h'_{\mathcal{G}}$ is a one-dimensional vector feature representation of molecular graph $\mathcal{G}$, and $h_{\mathcal{G}} \in [0, +\infty]$ is a scalar feature representation. Note that the notations $h_{\mathcal{G}}$ and $D(\bm{V}^{gen}, \bm{A}^{gen})$/$D(\bm{V}^{exist}, \bm{A}^{exist})$ have the same physical meaning, i.e., the output of the discriminator given a molecular graph $\mathcal{G}$ represented by $\left< \bm{V}^{gen},\bm{A}^{gen} \right>$ or $\left< \bm{V}^{exist},\bm{A}^{exist} \right>$.


\begin{figure*}[!htb]
	\centering	
	\includegraphics[width=1.0\textwidth]{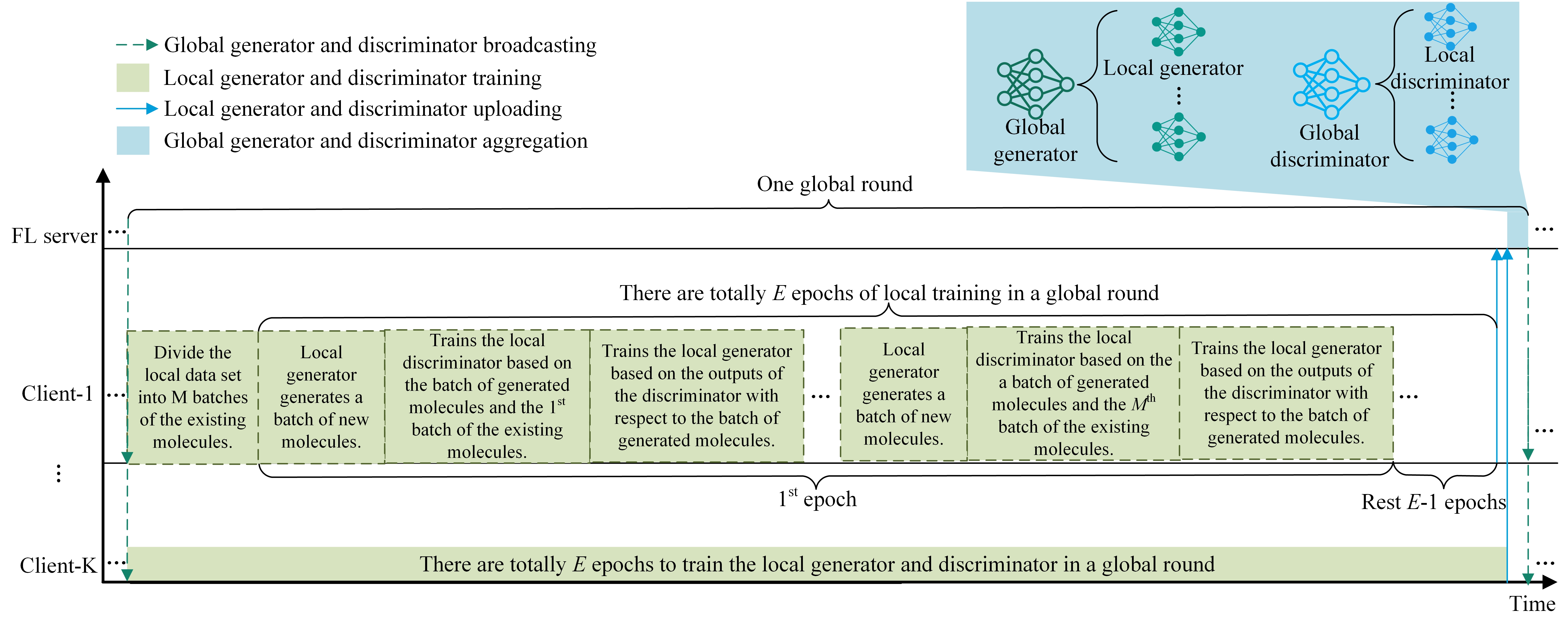} 
	\caption{Procedure of a global round in FL.}
	\label{fig:train_proc} 
\end{figure*}

\textbf{The FL system} enables distributed training of the generator and discriminator based on the local datasets in different clients in terms of pharmaceutical companies, ensuring that these clients do not need to share their data sets with others and thus preserving data privacy. Fig. \ref{fig:over} illustrates how the FL system distributively trains the generator and discriminator. In general, the system consists of a centralized FL server and a number of clients. Similar to the traditional FL system, there are four steps in each global round. Owing to the fact that there are two models, i.e., a generator and discriminator in GraphGANFed, the local model calculation step is different from the traditional FL. Specifically, 
\begin{enumerate}
\item \textbf{Global model broadcasting:} at the beginning of each global round, the FL server broadcasts the current generator and discriminator to the clients.
\item \textbf{Local model calculation:} upon receiving the global generator and discriminator, as shown in Fig.~\ref{fig:train_proc}, each client i) divides its local dataset of existing molecules into a number of batches with the same size, where $M$ is the number of batches, and ii) trains the local discriminator and generator over $E$ epochs. In each epoch, a client
\begin{enumerate}
    \item generates a batch of new molecules based on its local generator;
    \item trains the local discriminator based on the batch of generated molecules and the $1^{st}$ batch of the existing molecules in $M$ via batched gradients decent. Note that the node label/adjacency matrices of both generated and existing molecules, $\bm{V}^{gen}$/$\bm{A}^{gen}$ and $\bm{V}^{exist}$/$\bm{A}^{exist}$ are discrete and non-differentiable, meaning that back-propagation cannot be performed. Hence, we apply a gradient estimator \cite{gumbel2016}, which converts the discrete non-differentiable samples in $\bm{V}^{gen}$/$\bm{A}^{gen}$ and $\bm{V}^{exist}$/$\bm{A}^{exist}$  into continuous differentiable samples based on the Gumbel-Softmax distribution to enable back-propagation to be performed;
    \item trains the local generator based on the outputs of the discriminator with respect to the batch of generated molecules in Step a);
    \item repeats Steps a)-c) to train the local generator and discriminator until the last batch of the existing molecules in $M$ has been used for training; 
\end{enumerate} 


\item \textbf{Local model uploading:} once a client trained its local generator and discriminator after $E-1$ epochs, it uploads its local generator and discriminator to the FL server.
\item \textbf{Local model aggregation:} once the FL server receives the local generators and discriminators from the clients, it derives the new global generator and discriminator by aggregating all the local generators and discriminators based on, for example, FedAvg \cite{fedavg2017}.
\end{enumerate}
The global round continues until the global generator and discriminator converge. 

\subsection{Metrics} We apply the same metrics as the MOSES \cite{moses2020} benchmark to evaluate the performance of the molecules generated by GraphGANFed. These metrics can be used to identify the performance and some common issues (such as mode collapse and overfitting) of the generative model in GraphGANFed. Specifically, below are the details of all the metrics used to evaluate the generated molecules.
\begin{itemize}
    \item \textbf{Validity} refers to the fraction of generated molecules that are valid. RDkit's molecular structure parser \cite{2013rdkit} is used to compute the validity, which measures whether the generated molecules maintain explicit chemical constraints, such as proper atom valency and bond consistency. The \textbf{Validity} value is expressed as a percentage in the range $[0,100]$.
    
    \item \textbf{Uniqueness} refers to the fraction of valid molecules that are unique. Low uniqueness means repetitive molecular generation, which may imply mode collapse, i.e., it produces a limited variety of molecules. The value of \textbf{Uniqueness} can be a percentage in the range $[0,100]$.
    
    \item \textbf{Novelty} refers to the fraction of valid molecules that are different from the ones in the existing dataset. Low novelty indicates that the generative model tens to overfit to the existing data. The value of \textbf{Novelty} can be a percentage in the range $[0,100]$.
    
    \item \textbf{Internal Diversity (IntDiv$_p$)} estimates the chemical diversity in the generated molecules and is computed as the ($p^{th}$) power of the mean of the Tanimoto similarity ($T$) between the fingerprints of all the pairs of molecules ($m1, m2$) in the generated molecules set ($G$), i.e.,
    \begin{equation}
    \small
    \mathrm{IntDiv_{p}(G)} = 1 -  \sqrt[p]{\frac{1}{|G|^2}\sum_{m_{1},m_{2}\in G}T(m_{1}, m_{2})^p}.
    \label{e:intdiv}
    \end{equation}
    \textbf{IntDiv$_p$} can be used to identify normal failure cases in the generative model, such as mode collapse. During mode collapse, the generative model keeps generating similar structures or a small variety of molecules, excluding some parts of the chemical space. A higher \textbf{IntDiv$_p$} means a higher diversity in the generated molecule set. The value of \textbf{IntDiv$_p$} can be $[0,1]$. 
    \item \textbf{Quantitative Estimation of Drug-likeliness (QED)} refers to how likely a molecule is a viable candidate for a drug. The value of \textbf{QED} can be $[0,1]$.
    \item \textbf{Octanol-water partition coefficient (LogP)} refers to a chemical's concentration in the octanol phase to its concentration in the water phase in a two-phase octanol/water system. \textbf{LogP} was computed based on RDKit's Crippen estimation. The value of \textbf{LogP} can be $[0,1]$.
    \item \textbf{Similarity to a nearest neighbor (SNN)} measures an average Tanimoto similarity $(m_{G}, m_{R})$ between fingerprints of a molecule $m_{G}$ from the generated set $G$ and its nearest neighbor molecule $m_{R}$ in the real dataset ${R}$. It is computed by \cite{moses2020}:
    \begin{equation}
    \small
    \mathrm{SNN{(G, R)}} = \frac{1}{|G|}\sum_{m_{G}\in R}\max_{m_{R}\in R}T(m_{G}, m_{R})
    \label{e:snn}
    \end{equation}
    Here, we use the Morgan (extended connectivity) fingerprints estimation using the RDKit library. First, we compute the Morgan fingerprints of the generated molecule and randomly sample a number of molecules from the existing dataset. Then, we compare the similarities between the fingerprints of the generated molecule and each of the fingerprints of the existing molecules and compute the average of the similarity scores.  The similarity metric value can be represented as a measure of precision, and if the generated molecules are distinct from the manifold of the existing dataset, the similarity to the nearest neighbor will be very low. The value of \textbf{Similarity} is in the range of $[0, 1]$.
\end{itemize}
Note that all metrics, except for the \textbf{Validity}, are computed on only the valid molecules from the generated molecule set.

\section{Simulation Setups and Results}
In this section, we conduct extensive simulations by using benchmark datasets to evaluate our proposed GraphGANFed framework since this is the first work to perform molecular generation in the federated setting towards optimizing our evaluation metrics.
\subsection{Benchmark datasets}
We use three data sets from MoleculeNet.~\cite{moleculenet2018}: ESOL, QM8, and QM9. These datasets have different properties in terms of physical chemistry and quantum mechanics. Table \ref{Table: 1} presents the details of these datasets.   
\begin{itemize}
    \item ESOL \cite{esol2004} is a small dataset comprising water solubility data for 1,128 compounds. It has been used to train models which directly predict solubility from chemical structures that are encoded in SMILES string. 
    \item QM8 \cite{elect2015} originated from a recent survey on representing quantum mechanical calculations of electronic fields and excited state energy of small molecules. There are 22,000 molecules, each of which comprises up to eight heavy atoms, that are collected using techniques, such as second-order approximate coupled-cluster (CC2) and time-dependent density functional theories (TDDFT).
    \item QM9 \cite{qm92012} is an extensive dataset that provides geometric, electronic, energetic, and thermodynamic properties for a section of the GDB-17 database. It contains 134,000 stable organic molecules with up to 9 heavy atoms. The molecules in the dataset are reproduced using density functional theory (B3LYP/6-31G (2df,p) based DFT).
\end{itemize}

\begin{table}[t]
\centering
\tabcolsep 2pt
\caption{Datasets used in our experiments.}
\label{Table: 1}
\begin{tabular}{|c|c|c|c|c|}
\hline
  \textbf{\small Category} &\textbf{\small Dataset} & \textbf{\small \#Compds.} & \textbf{\small Avg. \#Atoms} & \textbf{\small Avg. \#Bonds} \\
  \hline
   {\small Physical Chem. } & {\footnotesize ESOL} & {\footnotesize 1,128} & {\footnotesize 13.29} & {\footnotesize 40.65} \\
 \hline
 {\small } & {\footnotesize QM8} & {\footnotesize 21,786} & {\footnotesize 7.77} & {\footnotesize 23.95} \\
 {\small Quantum Mech.} & {\footnotesize QM9} & {\footnotesize 133,885} & {\footnotesize 8.80} & {\footnotesize 27.60} \\
 \hline
 \end{tabular}
\end{table}

We randomly split each dataset into training, validation, and testing sets based on a ratio of 80:10:10. The training set is used to train the models via FL settings, and the validation and test sets are used for hyperparameter tuning and model evaluation, respectively. We distribute the samples in the training dataset to the clients based on the independent and identical distribution (IID) and non-IID, and analyze the performance of GraphGANFed under these two scenarios. Specifically, each sample in terms of a molecule is labeled by its molecular formula (e.g., $\mathrm{CH_3, CH_4, NH_3}$), and molecules with the same label are in the same class. For IID, molecules in each class are divided into $K$ groups of equal size. Each group is assigned to a client and $K$ is the total number of clients in FL. For non-IID, a random number of molecules will be selected in each class and allocated to each client.   

\subsection{Hardware and hyperparameter configurations}
In all simulations, we use a fixed $16$-dimensional random noise vector $\bm{z}$ sampled after every $1,000$ local epochs for ESOL and QM8, and $100$ local epochs for QM9. The batch size is $16$ and the number of local epochs is $E=1,000$ for each participated client. The gradient penalty coefficient is $\gamma= 10$ to calculate the loss function of the discriminator in Eq. \eqref{eq:loss_disc}. Also, both the generator and discriminator models are trained by using the Adam optimizer with the exponential decay rates $\beta_{1}=0.5$ and $\beta_{2}=0.999$, and the learning rate (LR) equals $0.0001$. Here, the LR for both generator and discriminator are decayed by a value of $100$ after each $1,000$ epochs. All the simulations are conducted on a computer with 2 NVIDIA Tesla K40m GPUs and 12 GB memory.

\subsection{Simulation results}
\subsubsection{Model convergence analysis}
\begin{figure}
\centering
    \includegraphics[width=\columnwidth]{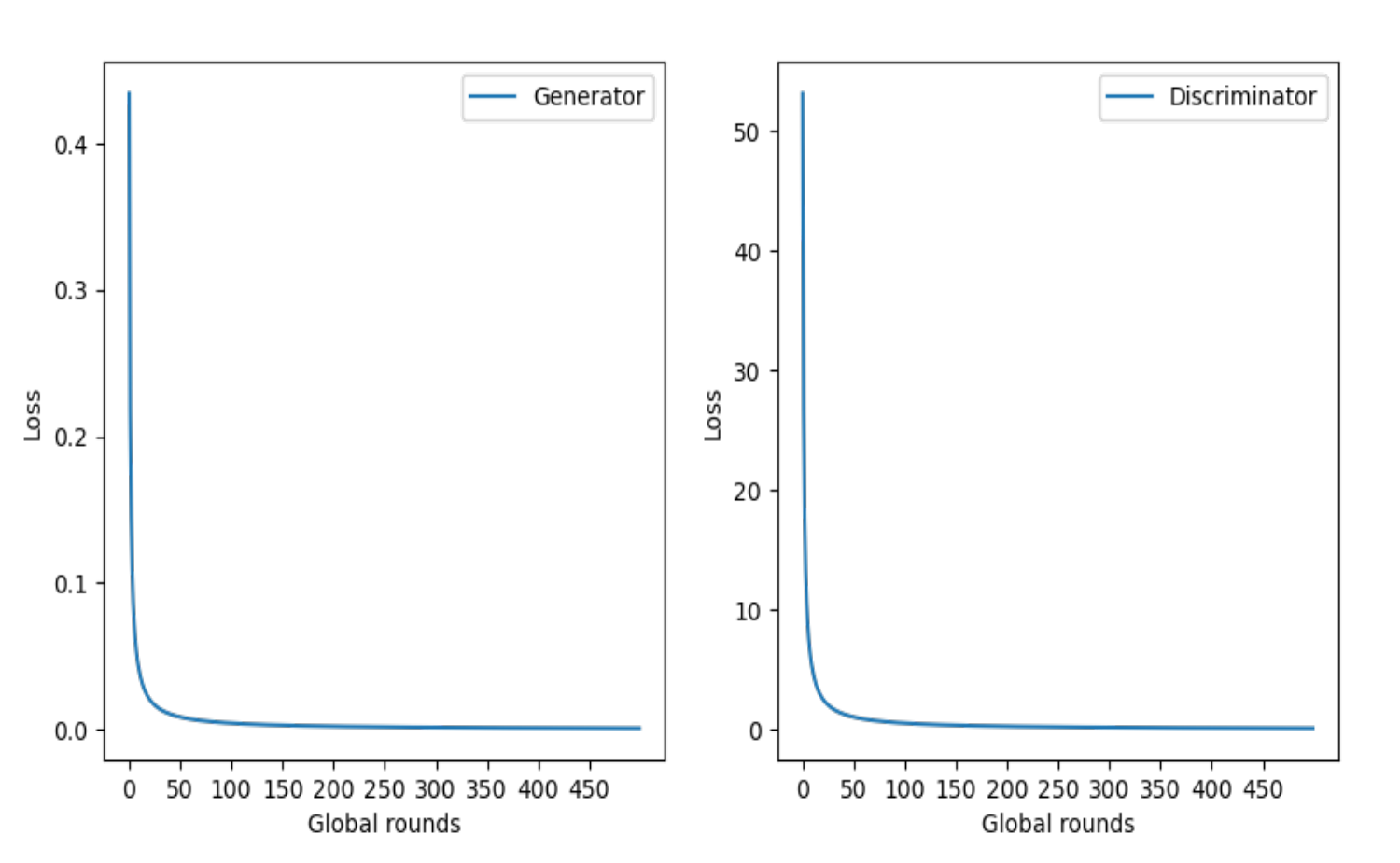}
    \includegraphics[width=\columnwidth]{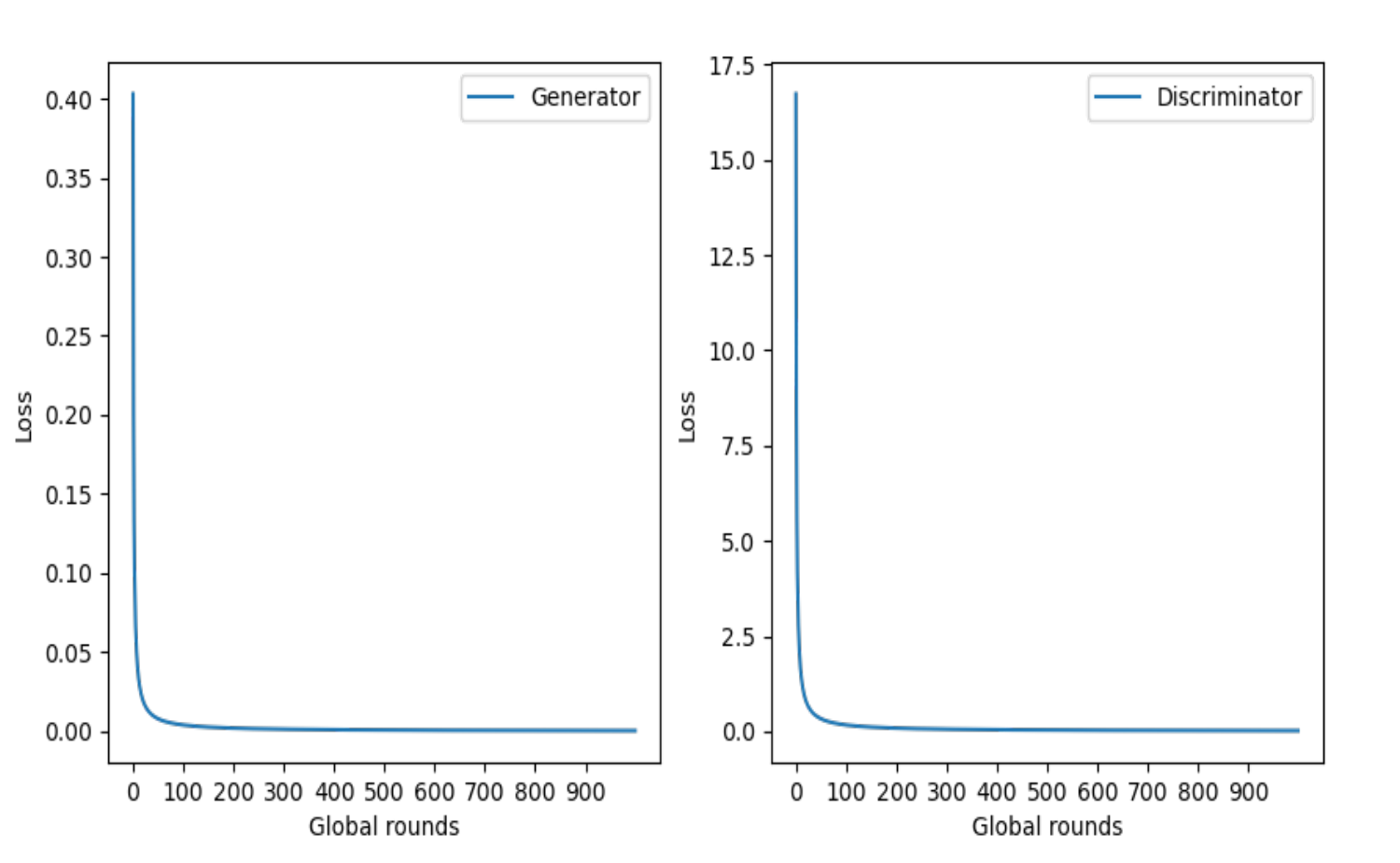}
\label{fig:loss_noniid}
\caption{Learning curves for the global generator and discriminator in IID (top) and non-IID (bottom).}
\label{fig:loss}
\end{figure}

Assume that the structures of the discriminator and generator are $([64, 128], 64, [128, 1])$ and $([32, 64, 128])$ respectively, where each value indicates the number of neurons/channels in a layer for the GCN model described in Fig. \ref{fig:over}. For example, $([64, 128], 64, [128, 1])$ implies that, in the discriminator, there are 64 and 128 channels in the first two convolution layers for atom feature extraction, 64 neurons in the hidden layer for feature dimension reduction, and 128 and 1 neurons in the last two layers for feature aggregation. There are $K=4$ clients to train the discriminator and generator based on FL, and Fig. \ref{fig:loss} shows the loss function curves of the discriminator and generator over different global rounds for the IID and non-IID settings. We can observe that both the generator and discriminator models converged after 60 global iterations and 150 global iterations in IID and non-IID settings, respectively, which demonstrates the feasibility of our proposed GraphGANFed framework. Also, it is common to have a slower convergence rate in the non-IID setting as compared to IID because of the high data distribution bias among the clients caused by non-IID.   


\subsubsection{Effect of different discriminator architectures on the metrics}
The discriminator is the key component to determine the performance of GraphGANFed, and so we will analyze the performance in terms of the values of the metrics (that we mentioned in Section III.C) by varying the complexity of the discriminator. Tables \ref{Table: 2} and \ref{Table: 3} show the results of applying different dimensions in terms of complexities of the discriminator for the IID and non-IID settings, respectively, where the second column in each table indicates different dimensions of the discriminator. 
The 3rd-8th columns imply the values of six metrics mentioned in Section III.C to evaluate the performance of the molecules generated by GraphGANFed. 

For ESOL, which is a small dataset consisting of only 1,128 compounds, we observe the results for the IID setting in Table \ref{Table: 2} and find that as the discriminator complexity increases, the values of QED, Uniqueness, LogP, and Similarity decrease, i.e.,  $\left( 0.54,0.34,0.34 \right)$, $\left( 100,0.9,0.9 \right)$, $\left( 0.73,0.26,0.26 \right)$, and $\left( 0.0146,0.0004,0.0004 \right)$, respectively, but the value of Validity increases, i.e., $\left( 70.5,100,100 \right)$. The same results are observed in the non-IID setting, as shown in Table \ref{Table: 3}. The reason for having the increasing Validity as the discriminator complexity increases is that training a higher complexity model over a small dataset, such as ESOL, leads to discriminator overfitting. This overfitting pushes the generator to generate molecules with the same or very similar structure as the molecules in the dataset, resulting in higher Validity. On the other hand, training a higher complexity model over a small dataset may lead to mode collapse, which reduces the values of QED, Uniqueness, and LogP. Thus, there is a tradeoff among Validity and QED/Uniqueness/LogP/Similarity, and the tradeoff can be adjusted by changing the discriminator complexity. In addition, for a small dataset like ESOL, a low discriminator complexity, i.e., [32, 64], 32, [64,1], is preferred to generate the molecules that have considerable values of all the metrics. 

\begin{table*}[!htb]
\centering
\tabcolsep 2pt
\caption{Performance of GraphGANFed with different discriminators and the same generator in IID.}
\label{Table: 2}
\begin{tabular}{|c|c|c|c|c|c|c|c|c|c|c|}
\hline
  \textbf{\footnotesize Datasets} & \textbf{\footnotesize Generator Dimension} & \textbf{\footnotesize Discriminator Dimension} & \textbf{\footnotesize Number of Clients} & \textbf{\footnotesize QED} & \textbf{\footnotesize Diversity} & \textbf{\footnotesize Validity} & \textbf{\footnotesize Uniqueness} & \textbf{\footnotesize Novelty} & \textbf{LogP} & \textbf{\footnotesize Similarity} \\
 \hline
 {\footnotesize} & {\footnotesize} & {\footnotesize [32,64],32,[64,1]} & {\footnotesize} & {\footnotesize 0.54} & {\footnotesize 1.00} & {\footnotesize 70.5} & {\footnotesize 100.0} & {\footnotesize 100.0} & {\footnotesize 0.73} & {\footnotesize 0.0146} \\
 {\footnotesize ESOL}  & {\footnotesize [32,128]} & {\footnotesize [32,128],32,[128,1]} & {\footnotesize 4} & {\footnotesize 0.34} & {\footnotesize 1.00} & {\footnotesize 100.0} & {\footnotesize 0.9} & {\footnotesize 100.0} & {\footnotesize 0.26}  & {\footnotesize 0.0004} \\
 {\footnotesize} & {\footnotesize} & {\footnotesize [32,256],32,[256,1]} & {\footnotesize} & {\footnotesize 0.34} & {\footnotesize 1.00} & {\footnotesize 100.0} & {\footnotesize 0.9} & {\footnotesize 100.0} & {\footnotesize 0.26}  & {\footnotesize 0.0004}\\
 \hline
  {\footnotesize } & {\footnotesize} & {\footnotesize [64,128],64,[128,1]} & {\footnotesize} & {\footnotesize 0.49} & {\footnotesize 1.00} & {\footnotesize 78.4} & {\footnotesize 18.1} & {\footnotesize 100.0} & {\footnotesize 0.54}  & {\footnotesize 0.0067} \\
 {\footnotesize QM8} & {\footnotesize[32,64,128]} & {\footnotesize [128,256],128,[256,1]} & {\footnotesize 4} & {\footnotesize 0.46} & {\footnotesize 0.99} & {\footnotesize 26.9} & {\footnotesize 51.4} & {\footnotesize 100.0} & {\footnotesize 0.59}  & {\footnotesize 0.0261}\\
 {\footnotesize} & {\footnotesize} & {\footnotesize [256,512],256,[512,1]} & {\footnotesize} & {\footnotesize 0.45} & {\footnotesize 0.99} & {\footnotesize 13.7} & {\footnotesize 67.9} & {\footnotesize 100.0} & {\footnotesize 0.62}  & {\footnotesize 0.0336} \\
 \hline
 {\footnotesize } & {\footnotesize} & {\footnotesize [64,128],64,[128,1]} & {\footnotesize} & {\footnotesize 0.45} & {\footnotesize 0.99} & {\footnotesize 17.5} & {\footnotesize 47.5} & {\footnotesize 100.0} & {\footnotesize 0.72}  & {\footnotesize 0.0004} \\
 {\footnotesize QM9} & {\footnotesize [64,128,256]} & {\footnotesize [128,256],128,[256,1]} & {\footnotesize 4} & {\footnotesize 0.46} & {\footnotesize 0.99} & {\footnotesize 32.2} & {\footnotesize 22.1} & {\footnotesize 100.0} & {\footnotesize 0.73}  & {\footnotesize 0.0004}\\
 {\footnotesize} & {\footnotesize} & {\footnotesize [256,512],256,[512,1]} & {\footnotesize} & {\footnotesize 0.50} & {\footnotesize 0.99} & {\footnotesize 72.9} & {\footnotesize 12.6} & {\footnotesize 100.0} & {\footnotesize 0.72}  & {\footnotesize 0.0056} \\
 \hline
 \end{tabular}
\end{table*}
\begin{table*}[hbt!]
\centering
\tabcolsep 2pt
\caption{ Performance of GraphGANFed with different discriminators and the same generator in non-IID.}
\label{Table: 3}
\begin{tabular}{|c|c|c|c|c|c|c|c|c|c|c|}
\hline
  \textbf{\footnotesize Datasets} & \textbf{\footnotesize Generator Dimension} & \textbf{\footnotesize Discriminator Dimension} & \textbf{\footnotesize Number of Clients} & \textbf{\footnotesize QED} & \textbf{\footnotesize Diversity} & \textbf{\footnotesize Validity} & \textbf{\footnotesize Uniqueness} & \textbf{\footnotesize Novelty} & \textbf{LogP}  & \textbf{\footnotesize Similarity} \\
 \hline
 {\footnotesize} & {\footnotesize} & {\footnotesize [32,64],32,[64,1]} & {\footnotesize} & {\footnotesize 0.45} & {\footnotesize 1.00} & {\footnotesize 55.4} & {\footnotesize 91.9} & {\footnotesize 100.0} & {\footnotesize 0.48}  & {\footnotesize 0.0015}\\
 {\footnotesize ESOL}  & {\footnotesize [32,128]} & {\footnotesize [32,128],32,[128,1]} & {\footnotesize 3} & {\footnotesize 0.34} & {\footnotesize 1.00} & {\footnotesize 100.0} & {\footnotesize 0.9} & {\footnotesize 100.0} & {\footnotesize 0.26}  & {\footnotesize 0.0004} \\
 {\footnotesize} & {\footnotesize} & {\footnotesize [32,256],32,[256,1]} & {\footnotesize} & {\footnotesize 0.34} & {\footnotesize 1.00} & {\footnotesize 100.0} & {\footnotesize 0.9} & {\footnotesize 100.0} & {\footnotesize 0.26}  & {\footnotesize 0.0004}\\
 \hline
  {\footnotesize } & {\footnotesize} & {\footnotesize [64,128],64,[128,1]} & {\footnotesize} & {\footnotesize 0.46} & {\footnotesize 0.99} & {\footnotesize 28.2} & {\footnotesize 54.7} & {\footnotesize 100.0} & {\footnotesize 0.43}  & {\footnotesize 0.0013}\\
 {\footnotesize QM8} & {\footnotesize[32,64,128]} & {\footnotesize [128,256],128,[256,1]} & {\footnotesize 3} &  {\footnotesize 0.49} & {\footnotesize 0.99} & {\footnotesize 66.3} & {\footnotesize 24.6} & {\footnotesize 100.0} & {\footnotesize 0.56}  & {\footnotesize 0.0053} \\
 {\footnotesize} & {\footnotesize} & {\footnotesize [256,512],256,[512,1]} & {\footnotesize } & {\footnotesize 0.49} & {\footnotesize 0.99} & {\footnotesize 88.9} & {\footnotesize 10.4} & {\footnotesize 100.0} & {\footnotesize 0.62}  & {\footnotesize 0.0378}\\
 \hline
 {\footnotesize } & {\footnotesize} & {\footnotesize [64,128],64,[128,1]} & {\footnotesize} & {\footnotesize 0.49} & {\footnotesize 0.99} & {\footnotesize 10.3} & {\footnotesize 75.0} & {\footnotesize 100.0} & {\footnotesize 0.46}  & {\footnotesize 0.0070}\\
 {\footnotesize QM9} & {\footnotesize[64,128,256]} & {\footnotesize [128,256],128,[256,1]} & {\footnotesize 3} &  {\footnotesize 0.50} & {\footnotesize 0.99} & {\footnotesize 56.2} & {\footnotesize 18.3} & {\footnotesize 100.0} & {\footnotesize 0.54}  & {\footnotesize 0.0073}\\
 {\footnotesize} & {\footnotesize} & {\footnotesize [256,512],256,[512,1]} & {\footnotesize} & {\footnotesize 0.50} & {\footnotesize 0.99} & {\footnotesize 73.5} & {\footnotesize 10.8} & {\footnotesize 100.0} & {\footnotesize 0.67}  & {\footnotesize 0.0078}\\
 \hline
 \end{tabular}
\end{table*}
For QM8, which is considered a medium dataset that has more compounds than ESOL but less than QM9, we observe the results for the IID setting in Table \ref{Table: 2} and find that as the discriminator complexity increases, the values of QED, Diversity, and Validity decrease, i.e., $\left( 0.49,0.46,0.45 \right)$, $\left( 1.00,0.99,0.99 \right)$, and $\left( 78.4,26.9,13.7 \right)$, respectively, but the values of Uniqueness, LogP, and similarity increases, i.e.,  $\left( 18.1,51.4,67.9 \right)$, $\left( 0.54,0.59,0.62 \right)$, and $\left( 0.0067,0.0261,0.0336 \right)$, respectively. The tradeoff changes from ``between Validity and QED/Uniqueness/LogP'' for ESOL into ``between QED/Diversity/Validity and Uniqueness/LogP/Similarity'' for QM8. Also, if we observe the results for non-IID in Table \ref{Table: 3}, we can find that as the discriminator complexity increases, only Uniqueness decreases, while QED, Validity, LogP, and Similarity increase. Hence, we conclude that the tradeoff among different evaluation metrics may change based on the size of the dataset and the data sample distribution among clients. In addition, for a medium dataset like QM8, a medium discriminator dimension, such as $[64,128],64,[128,1]$ or slightly smaller, is preferred to generate the molecules that have considerable metric values.   

For QM9, which is considered a large dataset, we can observe from the results that as we increase the discriminator complexity, QED, Validity, and Similarity increase, but Uniqueness reduces in the IID setting as shown in Table \ref{Table: 2}. In the non-IID setting, as shown in Table \ref{Table: 3}, only Uniqueness decreases, but QED, Validity, LogP, and Similarity increase as we increase the discriminator dimension. Hence, we can derive the same conclusion that the tradeoff among different metrics may change based on the size of the data set and the data sample distribution among clients. In addition, for a large dataset like QM9, a large discriminator complexity is preferred to generate the molecules that have considerable metric values.   

\begin{table*}[t!]
\centering
\tabcolsep 2pt
\caption{Performance of GraphGANFed with different numbers of clients in IID.}
\label{Table: 4}
\begin{tabular}{|c|c|c|c|c|c|c|c|c|c|c|}
\hline
  \textbf{\footnotesize Datasets} & 
  \textbf{\footnotesize Generator Dimension} & 
  \textbf{\footnotesize Discriminator Dimension} & 
  \textbf{\footnotesize Number of Clients} &  \textbf{\footnotesize QED} & \textbf{\footnotesize Diversity} & \textbf{\footnotesize Validity} & \textbf{\footnotesize Uniqueness} & \textbf{\footnotesize Novelty} & \textbf{LogP} & \textbf{\footnotesize Similarity} \\
 \hline
 {\footnotesize} & {\footnotesize} & {\footnotesize } & {\footnotesize 1} & {\footnotesize 0.46} & {\footnotesize 1.00} & {\footnotesize 91.9} & {\footnotesize 80.6} & {\footnotesize 100.0} & {\footnotesize 0.95} & {\footnotesize 0.0007} \\
 {\footnotesize ESOL}  & {\footnotesize [32,128]} & {\footnotesize[32,64],32,[64,1]} & {\footnotesize 5} & {\footnotesize 0.43} & {\footnotesize 1.00} & {\footnotesize 4.5} & {\footnotesize 100.0} & {\footnotesize 100.0} & {\footnotesize 0.88}  & {\footnotesize 0.0202} \\
 {\footnotesize} & {\footnotesize} & {\footnotesize } & {\footnotesize 10} & {\footnotesize 0.15} & {\footnotesize 1.00} & {\footnotesize 1.8} & {\footnotesize 100.0} & {\footnotesize 100.0} & {\footnotesize 1.00}  & {\footnotesize 0.0453}\\
 \hline
  {\footnotesize } & {\footnotesize} & {\footnotesize } & {\footnotesize 1} & {\footnotesize 0.45} & {\footnotesize 0.99} & {\footnotesize 21.9} & {\footnotesize 59.9} & {\footnotesize 100.0} & {\footnotesize 0.64}  & {\footnotesize 0.0202} \\
 {\footnotesize QM8} & {\footnotesize[32,64,128]} & {\footnotesize [64,128],64,[128,1]} & {\footnotesize 5} & {\footnotesize 0.44} & {\footnotesize 0.99} & {\footnotesize 4.6} & {\footnotesize 84.0} & {\footnotesize 100.0} & {\footnotesize 0.64}  & {\footnotesize 0.0281}\\
 {\footnotesize} & {\footnotesize} & {\footnotesize } & {\footnotesize 10} & {\footnotesize 0.44} & {\footnotesize 0.99} & {\footnotesize 4.6} & {\footnotesize 99.1} & {\footnotesize 100.0} & {\footnotesize 0.56}  & {\footnotesize 0.0108} \\
 \hline
 {\footnotesize } & {\footnotesize} & {\footnotesize } & {\footnotesize 1} & {\footnotesize 0.51} & {\footnotesize 1.00} & {\footnotesize 90.2} & {\footnotesize 3.9} & {\footnotesize 100.0} & {\footnotesize 0.71}  & {\footnotesize 0.0071} \\
 {\footnotesize QM9} & {\footnotesize[128,256,512]} & {\footnotesize [128,128],256,128,1} & {\footnotesize 5} & {\footnotesize 0.45} & {\footnotesize 1.00} & {\footnotesize 87.9} & {\footnotesize 11.9} & {\footnotesize 100.0} & {\footnotesize 0.34}  & {\footnotesize 0.0068}\\
 {\footnotesize} & {\footnotesize} & {\footnotesize } & {\footnotesize 10} & {\footnotesize 0.41} & {\footnotesize 1.00} & {\footnotesize 75.4} & {\footnotesize 13.9} & {\footnotesize 100.0} & {\footnotesize 0.24}  & {\footnotesize 0.0064} \\
 \hline
 \end{tabular}
\end{table*}

\begin{table*}[t!]
\centering
\tabcolsep 2pt
\caption{Performance of GraphGANFed with different discriminators and the same generator in non-IID.}
\label{Table: 5}
\begin{tabular}{|c|c|c|c|c|c|c|c|c|c|c|}
\hline
  \textbf{\footnotesize Datasets} & \textbf{\footnotesize Generator Dimension} & \textbf{\footnotesize Discriminator Dimension} &
  \textbf{\footnotesize Number of Clients} &
  \textbf{\footnotesize QED} & \textbf{\footnotesize Diversity} & \textbf{\footnotesize Validity} & \textbf{\footnotesize Uniqueness} & \textbf{\footnotesize Novelty} & \textbf{LogP} & \textbf{\footnotesize Similarity} \\
 \hline
 {\footnotesize} & {\footnotesize } & {\footnotesize } & {\footnotesize 1} & {\footnotesize 0.41} & {\footnotesize 1.00} & {\footnotesize 69.6} & {\footnotesize 74.4} & {\footnotesize 100.0} & {\footnotesize 1.00} & {\footnotesize 0.0047} \\
 {\footnotesize ESOL} & {\footnotesize [32,128]} & {\footnotesize [32,64],32,[64,1]} & {\footnotesize 3} & {\footnotesize 0.48} & {\footnotesize 1.00} & {\footnotesize 78.6} & {\footnotesize 87.5} & {\footnotesize 100.0} & {\footnotesize 0.48}  & {\footnotesize 0.0021} \\
 {\footnotesize} & {\footnotesize } & {\footnotesize } & {\footnotesize 7} & {\footnotesize 0.34} & {\footnotesize 1.00} & {\footnotesize 100.0} & {\footnotesize 0.9} & {\footnotesize 100.0} & {\footnotesize 0.26}  & {\footnotesize 0.0011}\\
 \hline
  {\footnotesize } & {\footnotesize } & {\footnotesize } & {\footnotesize 1} & {\footnotesize 0.46} & {\footnotesize 0.99} & {\footnotesize 48.9} & {\footnotesize 37.9} & {\footnotesize 100.0} & {\footnotesize 0.68} & {\footnotesize 0.0103} \\
 {\footnotesize QM8} & {\footnotesize [32,64,128]} & {\footnotesize [64,128],64,[128,1]} & {\footnotesize 3} & {\footnotesize 0.46} & {\footnotesize 0.99} & {\footnotesize 35.6} & {\footnotesize 47.5} & {\footnotesize 100.0} & {\footnotesize 0.66}  & {\footnotesize 0.0108}\\
 {\footnotesize} & {\footnotesize } & {\footnotesize } & {\footnotesize 7} & {\footnotesize 0.44} & {\footnotesize 0.99} & {\footnotesize 20.9} & {\footnotesize 59.3} & {\footnotesize 100.0} & {\footnotesize 0.65}  & {\footnotesize 0.0309} \\
 \hline
 {\footnotesize } & {\footnotesize } & {\footnotesize } & {\footnotesize 1} & {\footnotesize 0.50} & {\footnotesize 0.99} & {\footnotesize 73.8} & {\footnotesize 10.9} & {\footnotesize 100.0} & {\footnotesize 0.64}  & {\footnotesize 0.0066} \\
 {\footnotesize QM9} & {\footnotesize [128,256,512]} & {\footnotesize [128,128],256,[128,1]} & {\footnotesize 3} & {\footnotesize 0.50} & {\footnotesize 0.99} & {\footnotesize 85.6} & {\footnotesize 4.8} & {\footnotesize 100.0} & {\footnotesize 0.48}  & {\footnotesize 0.0078}\\
 {\footnotesize} & {\footnotesize } & {\footnotesize } & {\footnotesize 7} & {\footnotesize 0.50} & {\footnotesize 0.99} & {\footnotesize 77.3} & {\footnotesize 7.1} & {\footnotesize 100.0} & {\footnotesize 0.42}  & {\footnotesize 0.0039} \\
 \hline
 \end{tabular}
\end{table*}

\subsubsection{Effect of different client settings on the metrics}
In this section, we evaluate how different numbers of clients affect the performance of GraphGANFed. Tables \ref{Table: 4} and \ref{Table: 5} show the metrics of GraphGANFed under different numbers of clients in IID and non-IID settings, respectively. For ESOL, we observed that as we increase the number of clients, QED, and Validity decrease, i.e. $\left(0.46, 0.43, 0.15\right)$ and $\left(91.9, 4.5, 1.8\right)$, respectively, but Uniqueness, LogP, and Similarity increase, i.e. $\left(80.6, 100, 100\right)$, $\left(0.95, 0.88, 1\right)$, and $\left(0.0007, 0.0202, 0.0453\right)$, respectively, for the IID setting. For non-IID, we observed that as we increase the number of clients, only Validity increases, i.e. $\left(69.6, 78.6, 100\right)$, whilst the remaining metrics in terms of QED, Uniqueness, LogP, and Similarity decrease, i.e. $\left(0.41, 0.48, 0.34\right)$, $\left(74.4, 87.5, 0.9\right)$, $\left(1, 0.48, 0.26\right)$ and $\left(0.0047, 0.0021, 0.0011\right)$, respectively. The reason for having decreasing Validity as the number of clients increases in IID is that the number of data samples allocated to a client is reduced as the number of clients increases, and training the models over fewer data samples may result in overfitting, which leads the generator to produce molecules with limited variety in terms of reducing Validity. Hence, we can observe that there is a mode collapse happening on the Validity metric. We observed similar results for the non-IID setting where there was mode collapse in terms of Uniqueness as the number of clients increase. Hence the tradeoff changes from ``between QED/Validity and Uniqueness/LogP/Similarity" in IID to ``between QED/Uniqueness/LogP/Similarity and Validity" in non-IID as the clients are increased for ESOL. 

For QM8, we observed that as the number of clients increases, QED, Validity, and LogP decrease, but Uniqueness increases in both IID and non-IID settings. However, the Similarity decrease for IID, i.e. $\left(0.0202, 0.0281, 0.0108\right)$ and increase for non-IID, i.e. $\left(0.0103, 0.0108, 0.0309\right)$, as the number of clients increases. Hence, we can derive that the tradeoff changes from ``between QED/Validity/LogP/Similarity and Uniqueness" in IID to ``between QED/Validity/LogP and Uniqueness/Similarity" in non-IID as the number of clients increases for QM8. 

In the IID setting for QM9, we observed that as the number of clients increases, QED, Validity, LogP, and Similarity decrease, i.e. $\left(0.51, 0.45, 0.41\right)$, $\left(90.2, 87.9, 75.4\right)$, $\left(0.71, 0.34, 0.24\right)$, and $\left(0.0071, 0.0068, 0.0064\right)$, respectively, and Uniqueness increases, i.e., $\left(3.9, 11.9, 13.9\right)$. However, in the non-IID setting, Validity increases, i.e. $\left(73.8, 85.6, 77.3\right)$, and Uniqueness, LogP, and Similarity decrease, i.e. $\left(10.9, 4.8, 7.1\right)$, $\left(0.64, 0.48, 0.42\right)$, and $\left(0.0066, 0.0078, 0.0039\right)$, respectively. Hence, the tradeoff changes from ``between QED/Validity/LogP/Similarity and Uniqueness`` in IID to ``between Uniqueness/LogP/Similarity and Validity" in non-IID as the number of clients increases. Therefore, we can derive that the tradeoff among different evaluation metrics may change by varying the number of clients in the system and the distributions of data samples among the clients. 

\begin{figure}
    \centering
    \captionsetup{justification=centering}
    \includegraphics[width=\columnwidth]{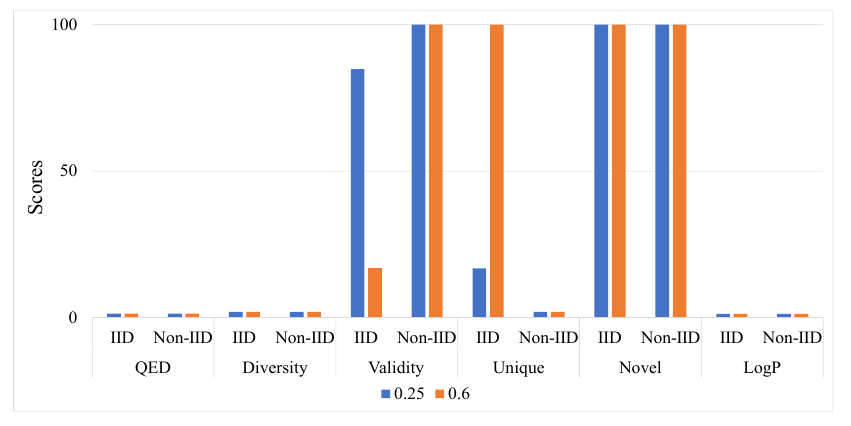}
    \caption{Performance of GraphGANFed by varying the dropout ratios for ESOL.}
    \label{fig:bar_esol}
\end{figure}

\begin{figure}
    \centering
    \captionsetup{justification=centering}
    \includegraphics[width=\columnwidth]{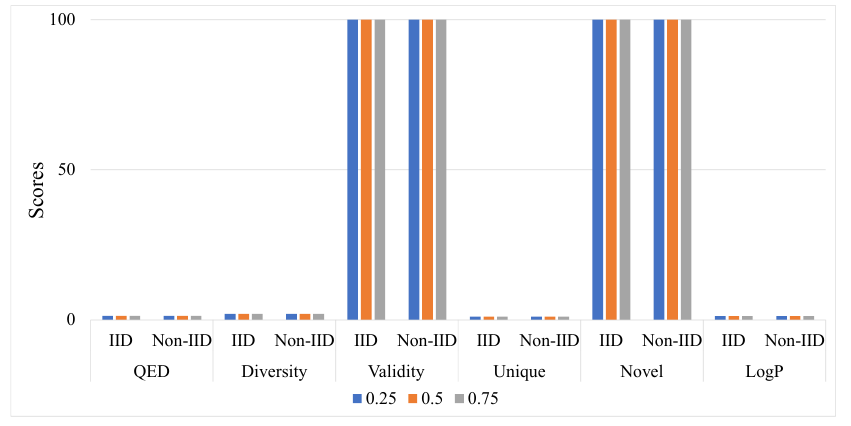}
    \caption{Performance of GraphGANFed by varying the dropout ratios for QM8.}
    \label{fig:bar_qm8}
\end{figure}

\begin{figure}
    \centering
    \captionsetup{justification=centering}
    \includegraphics[width=\columnwidth]{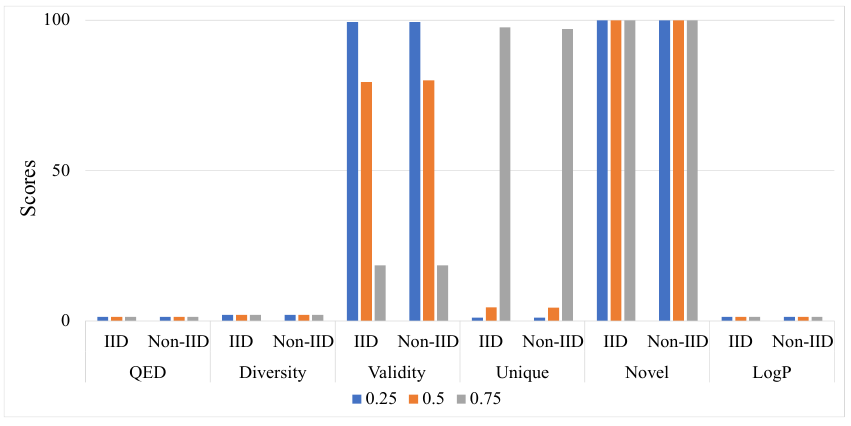}
    \caption{Performance of GraphGANFed by varying the dropout ratios for QM9.}
    \label{fig:bar_qm9}
\end{figure}

\subsubsection{Effect of different dropout ratios on generation metrics}
Applying dropout layers can reduce the model complexity without significantly affecting the model performance. We then analyze the performance of GraphGANFed if dropout layers are applied to both the generator and discriminator models. For the generator, we apply dropout after the last layer. For the discriminator, we apply dropout after the first two convolution layers and the last two parallel hidden layers. Assume that the structure of the generator is $\left[32, 128\right]$ for ESOL, $\left[32, 64, 128\right]$ for QM8, and $\left[64, 128, 256\right]$ for QM9, and the structure of the discriminator is $\left[32, 64],32,[64,1\right]$ for ESOL, $\left[64, 128],64,[128, 1\right]$ for QM8, and $\left[256,512],256,[512, 1\right]$ for QM9. The results of different dropout ratios are shown in Figs. \ref{fig:bar_esol}-\ref{fig:bar_qm9}.

Fig. \ref{fig:bar_esol} illustrates how different metrics change when the dropout ratio varies for ESOL. We can observe that as the dropout ratio increases from $0.25$ to $0.6$ in IID, Validity decreases from $84.8$ to $16.9$, and QED, Uniqueness, and LogP increase from $0.34$ to $0.35$, from $16.8$ to $100$, and from $0.26$ to $0.27$, respectively. For IID, the data distributions for the clients are the same. As the dropout ratio increases, more neurons are dropped in the generator and discriminator to reduce model complexity, but the models may not be able to learn some molecular representations in the dataset to generate valid samples, thus leading to a decrease in Validity. For non-IID, we observe that, as the dropout ratio increases from $0.25$ to $0.6$, the values of QED, Diversity, Validity, Uniqueness, Novelty, and LogP do not change. However, mode collapse happens on Uniqueness. Since it has been demonstrated that the high complexity of the discriminator lead to mode collapse on Uniqueness for ESOL  in Table \ref{Table: 2}, we can speculate that the complexity of the discriminator is still too high even if the  dropout ratio is $0.6$, thus leading to mode collapse. We also observe that even though the validity is $100\%$, the generator could only generate padding symbols $\left(*\right)$. 

Fig. \ref{fig:bar_qm8} shows how different metrics change when the dropout ratio varies for QM8. We observe that increasing the dropout ratios does not affect the metrics for both IID and non-IID. 
In addition, we also observe mode collapse on Uniqueness, which may be caused by the high model complexity even if the dropout ratio is $0.75$. Similar to the non-IID setting for ESOL, even though the Validity is $100\%$, the generator could only generate padding symbols $\left(*\right)$. 

Fig. \ref{fig:bar_qm9} shows how different metrics change when the dropout ratio varies for QM9. For IID, we can observe that as the dropout ratio increases from $0.25$ to $0.5$ to $0.75$, QED and Uniqueness increase, i.e. $\left(0.34, 0.34, 0.36\right)$ and $\left(0.1, 3.5, 97.6\right)$, respectively. Yet, Validity decreases, i.e. $\left(99.4, 79.4, 18.4\right)$. Similar results are observed in the non-IID setting that, as the dropout ratio increases,  QED and Uniqueness increase, but Validity decreases. 
Diversity and Novelty remain the same for all the dropout ratios in both IID and non-IID settings. 

Based on the observations in Figs. \ref{fig:bar_esol}, \ref{fig:bar_qm8} and \ref{fig:bar_qm9}, we can conclude that the model complexity can be adjusted by varying the dropout ratio. Also, having the right dropout ratio can avoid mode collapse. In addition, from all the experimental results presented above, we can observe that GraphGANFed can generate molecules to ensure high novelty $\left(\approx 100 \right)$ and diversity $\left(> 0.9\right)$.

\section{Conclusion}
In this paper, we have proposed the GraphGANFed framework, which integrates GCN, GAN and FL to generate novel and effective molecules for drug discovery, while preserving privacy. In GraphGANFed, there is an FL server, which aims to train a global generator and a global discriminator in a distributed manner, where the global generator tries to generate new molecules that can preserve the properties of the existing molecules to deceive the global discriminator and the global discriminator is to distinguish generated molecules from the existing molecules. In each global iteration, an FL server enables different clients to participate in the training process to train their local generators and discriminators over local datasets. The local generators and discriminators are uploaded and aggregated to derive a new global generator and global discriminator for the next global iteration. Extensive simulations have been made to prove the feasibility and effectiveness of GraphGANFed. The molecules generated by GraphGANFed can ensure high novelty $\left(\approx 100 \right)$ and diversity $\left(> 0.9\right)$. Also, the simulation results suggest that 1) a lower complexity discriminator model can better avoid mode collapse for a smaller dataset in IID and non-IID, 2) there is a tradeoff among different metric values and the tradeoff may change based on the size of the dataset and the data sample distribution, 3) model complexity of the generator and discriminator can be adjusted by varying the dropout ratio, and having the right dropout ratio can avoid mode collapse. 

\bibliographystyle{IEEEtran}
{\footnotesize \bibliography{IEEEabrv,ref}}
\end{document}